\documentclass[twocolumn,10pt]{asme2e}

\usepackage{graphicx}
\usepackage{appendix}
\usepackage{adjustbox}
\usepackage{amsmath}
\usepackage{flafter} 
\usepackage{multirow}
\usepackage{rotating}
\usepackage{amsmath}
\usepackage{amssymb}
\usepackage{subfigure}
\usepackage{hyperref}
\usepackage{color}
\usepackage{booktabs}
\usepackage{graphicx}
\usepackage{dblfloatfix}
\usepackage{relsize}
\usepackage{multirow} % allows multirow in tables
 \usepackage{array}
 \usepackage{float}
 \usepackage{longtable}
 \usepackage{balance}
\usepackage{vcell}
\usepackage{graphicx}
\usepackage{makecell}
\usepackage[table,xcdraw]{xcolor}
\usepackage{rotating}
\usepackage{pifont}
\newcommand{\multirot}[1]{\multirow{2}{*}[1.5ex]{\rotcell{\rlap{#1}}}}

\def\code#1{\texttt{#1}}

\confshortname{IDETC/CIE 2022}
\conffullname{the ASME 2022 \\International Design Engineering Technical Conferences and \\Computers and Information in Engineering Conference}

\confdate{August 14-17}
\confyear{2022}
\confcity{St. Louis}
\confcountry{Missouri}

\papernum{IDETC-CIE 2022/DAC-88049}

\title{MATERIAL PREDICTION FOR DESIGN AUTOMATION USING GRAPH REPRESENTATION LEARNING}

\author{Shijie Bian
    \affiliation{
	Department of Mathematics\\
	University of California\\
	Los Angeles, CA 90095, USA
    }	
}

\author{Daniele Grandi, Kaveh Hassani
    \affiliation{
	Autodesk Research\\
	Autodesk Inc.\\
    San Rafael, CA 94903, USA
    }	
}

\author{Elliot Sadler, Bodia Borijin, \\ \textbf{Axel Fernandes}
    \affiliation{Autonomy Research Center for STEAHM\\
	California State University Northridge\\
	Northridge, CA 91324, USA
    }
}

\author{Andrew Wang
    \affiliation{Portola High School\\
	Irvine, CA 92618, USA\\
    }
}

\author{Thomas Lu, Richard Otis
    \affiliation{Jet Propulsion Laboratory\\
    California Institute of Technology\\
	Pasadena, CA 91109, USA\\
	}
}

\author{Nhut Ho, Bingbing Li\thanks{Address all correspondence to this author.}
    \affiliation{Autonomy Research Center for STEAHM\\
	California State University Northridge\\
	Northridge, CA 91324, USA\\
	*bingbing.li@csun.edu
    }
}

\begin{document}

\maketitle    

%%%%%%%%%%%%%%%%%%%%%%%%%%%%%%%%%%%%%%%%%%%%%%%%%%%%%%%%%%%%%%%%%%%%%%
\begin{abstract}
{\it Successful material selection is critical in designing and manufacturing products for design automation. Designers leverage 
their knowledge and experience to create high-quality designs by selecting the most appropriate materials through performance, manufacturability, and sustainability evaluation. Intelligent tools can help designers with varying expertise by providing recommendations learned from prior designs. To enable this, we introduce a graph representation learning framework that supports the material prediction of bodies in assemblies. We formulate the material selection task as a node-level prediction task over the assembly graph representation of CAD models and tackle it using Graph Neural Networks (GNNs). Evaluations over three experimental protocols performed on the Fusion 360 Gallery dataset indicate the feasibility of our approach, achieving a 0.75 top-3 micro-$F_1$ score. 
The proposed framework can scale to large datasets and incorporate designers' knowledge into the learning process. 
These capabilities allow the framework to serve as a recommendation system for design automation and a baseline for future work, narrowing the gap between human designers and intelligent design agents.}
\end{abstract}

%%%%%%%%%%%%%%%%%%%%%%%%%%%%%%%%%%%%%%%%%%%%%%%%%%%%%%%%%%%%%%%%%%%%%%
% \begin{nomenclature}
% \entry{A}{You may include nomenclature here.}
% \entry{$\alpha$}{There are two arguments for each entry of the nomemclature environment, the symbol and the definition.}
% \end{nomenclature}

% The spacing between abstract and the text heading is two line spaces.  The primary text heading is  boldface in all capitals, flushed left with the left margin.  The spacing between the  text and the heading is also two line spaces.

%%%%%%%%%%%%%%%%%%%%%%%%%%%%%%%%%%%%%%%%%%%%%%%%%%%%%%%%%%%%%%%%%%%%%%
\section{INTRODUCTION}

\begin{figure*}[h!]
\begin{centering}
\includegraphics[width=19cm]{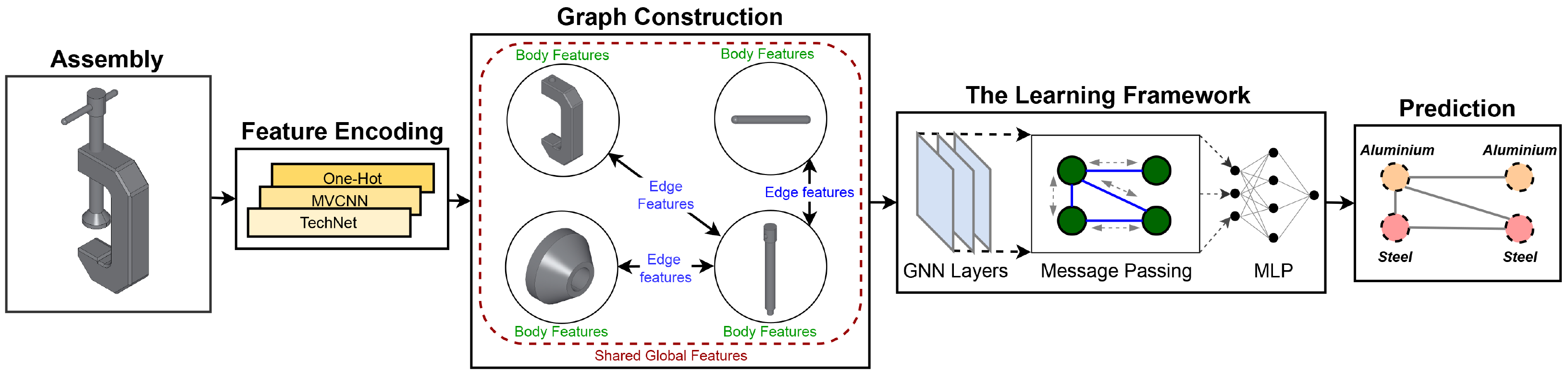}
\end{centering}
\caption{ \footnotesize PROPOSED FRAMEWORK FOR PREDICTING MATERIALS OF ASSEMBLY BODIES USING GRAPH REPRESENTATION LEARNING.}
\label{fig:framework} 
\end{figure*}

Material selection is a critical step in designing and manufacturing a product. Materials are integral to a product's success, safety, and efficiency \cite{arnold2012materials}. Engineers and designers spend a considerable amount of time evaluating trade-offs between materials, such as cost, performance, manufacturability, and sustainability \cite{mouritz2012introduction}. Each material needs to support the product's primary function while also withstanding uncertain environmental conditions that the product undergoes. Selecting an improper material can lead to extra expense, low structural integrity, and a shortened product life cycle \cite{10.31399/asm.hb.v11.a0003501}. Designers mainly rely on their experience and existing guidelines to evaluate trade-offs between different materials. Designing intelligent tools based on knowledge graphs and large design data could help facilitate this process. In product design, efforts to consolidate design knowledge in knowledge graphs have resulted in robust graph representations of domain-specific semantic relationships\cite{Sarica2020, ferrero2022classifying}. Engineering design knowledge graphs are useful for concept generation and evaluation \cite{siddharth2022engineering}. However, there is an opportunity to expand on prior work towards later phases of the design process by extracting design knowledge from computer-aided design (CAD) models to complement existing semantic networks.

CAD tools are used to create 2D and 3D models of the products and document various aspects of the design, including the geometries, dimensions, tolerances, degrees of freedom, and relative motions of parts, as well as the material of each component in the assembly \cite{sargent2016materials}. The material information stored in CAD tools is typically used for simulation or rendering workflows. Both activities help designers assess and visualize trade-offs between different materials and identify the best materials for the design. 

In recent years, several large datasets of curated CAD models have been released that support machine learning methods for various data-driven design applications \cite{willis2020fusion, Koch_2019_CVPR, sangpil2020large}. Simultaneously, graphs have been leveraged to represent design data and to capture various relationships, including joint relationships between parts in assemblies \cite{willis2021joinable, jones2021sb}, semantic relationships between engineering and design concepts \cite{Sarica2020}, and geometric relationships between faces of BREPs \cite{lambourne2021brepnet, jayaraman2020uv}.

In product design, efforts to consolidate design knowledge in knowledge graphs have resulted in robust graph representations of domain-specific semantic relationships. Engineering design knowledge graphs have been shown to be useful for concept generation and evaluation \cite{ferrero2022classifying}.Leveraging the expressive power of graphs in capturing multi-modal information, we propose a framework (Figure~\ref{fig:framework}) in which material prediction is posed as a node prediction task and is tackled through learning representations using graph neural networks (GNNs). Motivated by the importance of material selection to support design automation, this unified framework can help designers select appropriate materials by providing part-level material suggestions given a product assembly. 

The method is validated through three experiments simulating different applications using the Fusion 360 Gallery Assembly Dataset \cite{willis2021joinable}. Also, through an ablation study of features and GNN layers, additional insights are provided into factors that influenced the performance, which may serve as a preliminary discussion and a baseline for future studies.

To summarize, our work makes the following contributions to the areas of design automation and engineering:
\begin{enumerate}
    \item We study the material selection task for design automation and devise a systematic procedure to represent CAD models as graphs by extracting and encoding multi-modal features.
    \item We leverage GNNs to learn expressive representations from existing CAD assemblies and use them for predicting materials on new assemblies.
    \item We provide insights into the importance of features commonly found in CAD with respect to material prediction and evaluate the framework within three design scenarios.
\end{enumerate}

To enable further research and reproducibility, we share the code for feature extraction, training, and experiments \footnotemark[1].

\footnotetext[1]{\url{https://github.com/danielegrandi-adsk/material-gnn}}

%%%%%%%%%%%%%%%%%%%%%%%%%%%%%%%%%%%%%%%%%%%%%%%%%%%%%%%%%%%%%%%%%%%%%%
\section{RELATED WORK}
We review the prior work related to material selection, graph representation learning in design automation, and graph neural networks. Reflecting on these past works, we provide deeper insights into the motivation and vision of our work.

\subsection{Material Selection Design}
In product design, material selection can be broken down into a five-step procedure: (1) establishing design requirements, (2) screening materials, (3) ranking materials, (4) researching material candidates, and (5) applying constraints to the selection process \cite{arnold2012materials}. Many properties influence material selection, and the needs of a project determine the most important factors to be considered. This step is heavily reliant on performance indices and material property charts (Ashby charts) \cite{arnold2012materials} which plot one material property against another, resulting in clusters of material classes. Then, the material search can be further narrowed down by setting axis limits \cite{doi:10.1179/mst.1989.5.6.517}. In practice, tools such as Granta CES Edupack software can be used to compare over 4,000 materials based on user-input design requirements and constraints \cite{ullah2020investigation}. Product design aims to meet customer needs of technical requirements \cite{ALBINANA2012433}. The material selection factors become more ambiguous because of the wide range of consumer products and multiple ways of designing each product. Material aspects such as quality, cost, and function must be considered in the design of user products because they interact and contribute to the final design \cite{van_kesteren_2008}. Another important approach focuses on sustainability as a consideration of the economic, environmental, and societal impacts during the whole life-cycle from cradle to stage\cite{zarandi2011material}. When working with complex assemblies, the material selection process is further complicated, as the material of individual parts affects the assembly integration process \cite{doi:10.1080/002075431000087265}. Thus, assembly data should be considered during the material selection process. 

\subsection{Graph Neural Networks (GNNs)}
\label{sec:GNN}
GNNs are a class of deep neural networks to perform inference on data decribed by graphs in a non-Euclidean space. GNNs learn node representations over order-invariant and variable-size data, structured as graphs, through an iterative process of transferring, transforming, and aggregating the representations from topological neighbors. The learned representations are then summarized into a graph-level representation \cite{li_2015_iclr, gilmer_2017_icml, kipf_2017_iclr, velickovic_2018_iclr, xu_2019_iclr, Khasahmadi2020Memory}. Utilizing the process of representation learning, GNNs can perform tasks such as graph, node, and edge classifications, and are applied to real-world applications such as point cloud segmentation \cite{hassani_2019_iccv}, robot designs \cite{wang_2018_iclr}, physical simulations\cite{sanchez_2018_icml, sanchez2020learning}, quantum chemistry \cite{gilmer2017neural}, material design \cite{guo2020semi}, semantic role labeling \cite{marcheggiani2017encoding}, and product relationship predictions \cite{ahmed2021graph}.

\subsection{Graph Representation of CAD Models}
The design process as a whole is iterative and generates large amounts of data that can be organized and parsed for additional information that may be used to improve the design \cite{doi:10.1080/00207543.2018.1443229}. This information, collected from all aspects of the product life-cycle, is multi-modal and can be in the form of semantic names, customer requirements, 3D geometry, material properties, manufacturing tolerances, cost information, etc. This data may then be used to modify the design itself, enabling some automation of the design process by learning from prior examples \cite{feng2020data, funkhouser2004modeling}. Prior work looked at organizing and learning from design knowledge acquired from sources such as taxonomy-based design repositories \cite{ferrero2022classifying, Williams2019}, product teardowns \cite{wang2021understanding}, patent data \cite{Sarica2020}, and geometry-based design repositories \cite{chaudhuri2020learning, li2020learning, mo2019structurenet}. When working with geometric data, graphs have been leveraged to represent CAD models with goals ranging from representation of complex relations to solving design problems \cite{slyadnev2020role, HUANG20081073, SUNIL2010686, lambourne2021brepnet, jayaraman2020uv}. The current work expands the representation of design into multi-modal representation, capturing geometric and semantic information in addition to the hierarchical structure of parts in assemblies. 

% The current work expands on prior literature by expanding the representation of the design with the use of multi-modal data, and concurrently represents various design features present in CAD, including the geometry, semantic name, and hierarchical relationships of parts in assemblies. 

% Graphs have been used to represent BREP geometries from CAD for feature recognition\cite{SUNIL2010686}, and for direct use with neural networks \cite{lambourne2021brepnet, jayaraman2020uv}. 

% For an overview of GNNs see \cite{zhang_2020_kde, wu_2020_nnls}.

% [Can reference the format of "Graph Neural Networks - Note" (by Nihal V. Nayak): https://nihalnayak.github.io/assets/notes/gnn_notes.pdf]
% Implementing GCN: https://pytorch-geometric.readthedocs.io/en/latest/notes/create_gnn.html

%%%%%%%%%%%%%%%%%%%%%%%%%%%%%%%%%%%%%%%%%%%%%%%%%%%%%%%%%%%%%%%%%%%%%%
\section{METHODOLOGY}

This section provides an overview of the framework, the process and motivation for data selection, and the detailed methodology for predicting materials of assembly bodies using graph representation learning. 

% \subsection{Framework}
As illustrated in Figure \ref{fig:framework}, the proposed framework consists of three main modules: (1) the \textbf{feature encoding} module extracts useful features from assembly files (such as semantic and physical properties of bodies, assembly connections between each body, and global features shared across bodies) and encodes them into embedding vectors; (2) the \textbf{graph construction} module structures assemblies, along with their bodies and the encoded features, into assembly graphs; (3) the \textbf{learning framework} trains GNNs upon the graph representations to learn robust embeddings that are used for predicting materials.

\subsection{Dataset}
\label{source-of-data}

\begin{figure}[t]
\begin{center}
\includegraphics[width=9cm]{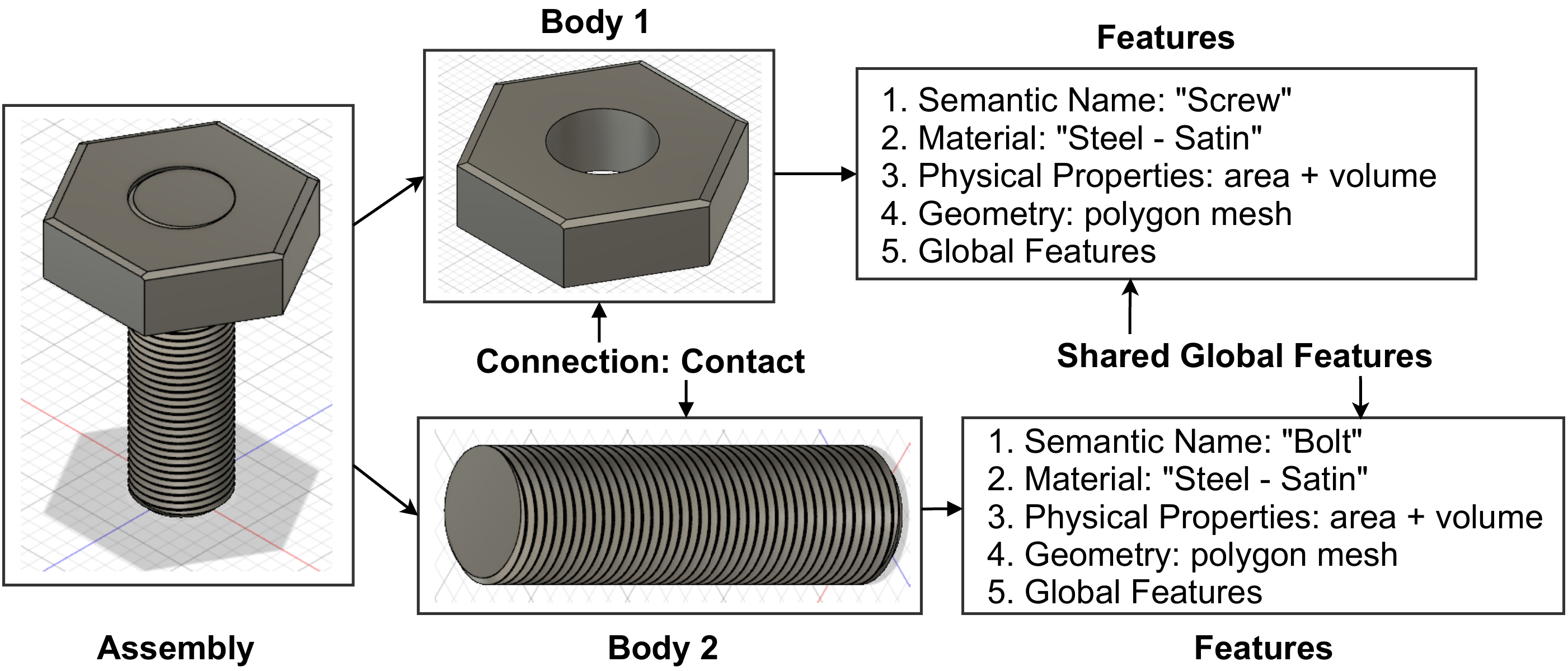}
\end{center}
\caption{\uppercase{\footnotesize AN EXAMPLE of an assembly with two bodies and their features.}}
\label{fig:feature_engineering} 
\end{figure} 

The proposed framework is developed to work with the publicly available Fusion 360 Gallery Assembly Dataset \cite{willis2021joinable}. As illustrated in Figure \ref{fig:feature_engineering}, each \textit{assembly} in the dataset contains several \textit{bodies} with a set of design features. Users organize \textit{bodies} by placing them in a hierarchy of \textit{occurrences}, which are the building blocks that make up \textit{assemblies}. The dataset contains CAD data from 154,468 bodies, grouped into 8,251 assemblies, from different industries and with various levels of detail. It was chosen because of its size, diversity of designs and features, ease of processing, and presence of per-body material labels. We use the train/test split provided with the dataset for all experiments. For details on features and the encoding methods refer to Table~\ref{table:assembly-features} in Appendix~\ref{appendix:feature-list}. We applied several pre-processing steps to the dataset as follows.

% The JSON files containing the design data, the OBJ files representing the part geometries, and the train-test split made available with the dataset were used for all experiments presented in this work. Table~\ref{table:assembly-features} in Appendix~\ref{appendix:feature-list} shows a list of the design features extracted from the dataset, a description of the features, as well as the methods used to encode the features as either edge or node feature in the graphs.

% \subsubsection{Data Pre-processing}

% After selecting the dataset for investigation, several pre-processing steps were applied to clean the data and help the learning architecture extract helpful information
 
\paragraph{Data Filtering} Many bodies in the dataset are labeled with the default material assigned at the time of creation with the CAD tool. Suppose all the bodies of an assembly are made of the default material, then it might be assumed that the user did not perform intentional material selection, rendering the data unusable for learning. Therefore, a filtering step was taken to remove the assemblies that have all bodies labeled with the default material and default appearance. 5388 out of 8251 assemblies were left after this step.

\paragraph{Material ID Transformation} Each body in the dataset has two material labels: a physical material label that defines the mechanical and physical properties of the body, and an appearance label used for rendering. While each label serves a different purpose, it can be assumed that the designer intends each body to be defined by only one material. Therefore, for bodies with default physical material but non-default appearance labels, the non-default appearance was used as the ground truth. This was done to improve the quality and number of ground-truth labels.

\paragraph{Material Grouping} The user-defined body materials are significantly diverse, skewed, and sparse in distribution, as shown in Table~\ref{table:material-count} in Appendix~\ref{appendix:dataset-statistics}. Considering that the materials with infrequent occurrences are unlikely to appear in future test cases and may not generalize well to other design use cases, only the top 20 material categories were preserved. Other materials were grouped into a single category.

\subsection{Feature Encoding}\label{sec:feature_engineering}

This section describes the definition and encoding of valuable features from the dataset. These encoded features are later used to generate relational assembly graphs in Section~\ref{sec:graph-construction} to feed into the learning architecture.

\begin{figure}[t]
\begin{center}
\includegraphics[width=9cm]{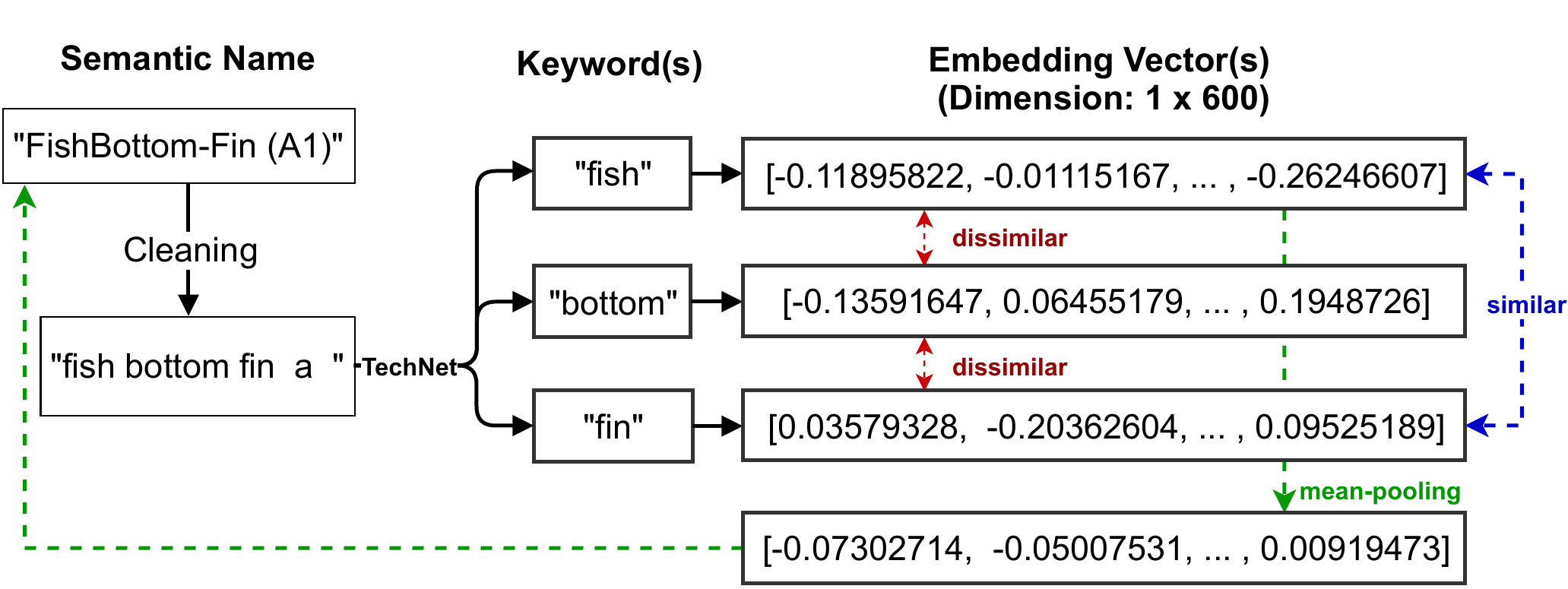}
\end{center}
\caption{AN EXAMPLE OF FEATURE ENCODING FOR THE SEMANTIC NAMES OF AN ASSEMBLY.}
\label{fig:semantic_embeddings} 
\end{figure}

\paragraph{Semantic Names}
Each body has a \code{body name} and an \code{occurrence name}. Users can manually set these semantic names or leave them as default. TechNet\cite{Sarica2020}, a semantic network with support for processing technology-focused texts, was used to pre-process and vectorize the semantic names. The cleaned semantic names were passed to TechNet to extract and retrieve keywords and corresponding 600-dimensional embeddings of keywords. The unmatched names were assigned a feature vector imputed according to the distribution of the TechNet database, while bodies with the default name were assigned a vector of zeros. The average embedding of all the keywords was assigned to the semantic name for semantic names with multiple keywords. The process is shown in Figure \ref{fig:semantic_embeddings}.

% To vectorize the semantic names we used TechNet\cite{Sarica2020}, a semantic network that supports the analysis and processing of technology-focused texts via a vectorized structure. The TechNet API was used to extract keywords from the cleaned semantic names and a corresponding $1 \times 600$ embedding vector was also retrieved. The names that remained unrecognized by TechNet after the pre-processing steps were assigned a random vector generated according to the distribution of the TechNet database. For semantic names with multiple detected keywords, their embedding was generated by taking a mean-pooling of the embedding of each of its extracted keywords. For ``root bodies", the embedding vector of their occurrence names are configured as a vector of zeros. A standard scalar transformation was then applied to the encoded semantic names to enable a better distribution for the learning architecture to take in.

% The final distributions of the semantic names after pre-processing are summarized in Figure~\ref{} in Appendix~\ref{appendix:dataset-statistics}.

\paragraph{Physical Properties}
Each body has three physical properties: \code{body area} (in square meters), \code{body volume} (in cubic meters), and \code{body center of mass} (represented in the \textit{x, y, z} coordinates). All physical properties are generated by Autodesk Fusion 360 during the user's design process. The physical properties are normalized to the standard scale.

\paragraph{Geometric Information}
The 3D geometry of every body is used to render 2D depth images from 12 different views equally spaced around the object. A Multi-View Convolutional Neural Network (MVCNN)~\cite{su15mvcnn} is trained by feeding it the 2D images and enforcing it to classify the geometric shapes into 21 classes selected for the experiments. Once trained, the model's weights are frozen, and the classification layer is truncated. The model is then used as a
pre-trained encoder to compute 512-dimensional visual embeddings of all the available geometric data. The visual embeddings are used to represent the \code{body geometry}. The MVCNN is trained using the same train-test split as the GNN.
% , allowing the geometric embeddings to be used as features in the assembly graphs to train the GNN with no data leaks between sets. This also allows us to establish a baseline method to compare against the performance of the GNN, as shown in Section~\ref{sec:general-setup}. 

\paragraph{Connection Types} \label{sec:connection_types}
Each assembly in the dataset comprises multiple interconnected bodies via various assembly relationships. Specifically, we consider three predominant types of connections. \code{Contacts} define the relationship between two bodies whose faces are in contact with each other. \code{Joints} define the relationship between two bodies whose relative pose and degrees of freedom are constrained. \code{Hierarchical} edges represent the user-assigned relative hierarchy of the bodies, indicating whether multiple bodies share the same occurrence. The connection types are encoded using the one-hot method as a categorical feature. % A histogram of the distribution of the connection types is shown in Figure~\ref{} of Appendix~\ref{appendix:dataset-statistics}.

\paragraph{Global Features}
Each assembly has features that are shared globally among the bodies of the same assembly. These features include the \code{assembly physical properties}, \code{assembly geometric properties}, \code{design category}, \code{industry},  and \code{products} used to create the design. For more details see Appendix \ref{appendix:feature-list}. 
% In general, global features represented as floating-point or integer numbers are preserved and later normalized using Standard Scalar transformation. Categorical features represented as character strings are one-hot encoded.

\subsection{Graph Construction}\label{sec:graph-construction}
% This section explains the method used to structure CADs as graphs. The features described in this section are extracted following the methodologies described in Appendix \ref{appendix:dataset-statistics} and encoded following the methodologies introduced in Section \ref{sec:feature_engineering}.
% \paragraph{Graph Overview} 
Each assembly is represented as a directed and attributed assembly multi-graph, where bodies that correspond to individual parts are represented as nodes. Connections that represent assembly relationships are represented as edges. The graph is directed since each edge contains a source node, a destination node, and a direction. The graph is attributed since each node and edge contains encoded features. Furthermore, the graph allows multiple edges between any pair of nodes. 

% An example assembly graph is shown in Figure X.

\paragraph{Graph Nodes} Each body is represented as a node containing semantic name embeddings, visual embeddings, and physical properties. The encoded global features of the graph are concatenated with the node feature and shared across the nodes of the same graph. The ground-truth material names encoded as one-hot vectors are also considered graph node attributes and used in the partial algorithm-guided experiment.

\paragraph{Graph Edges} The body nodes are connected through edges that represent assembly relationships between the bodies (Section \ref{sec:connection_types}). Specifically, \code{Assembly Edge} contains the one-hot encoding of connection type. Note that there can be multiple edges between a pair of nodes, corresponding to the multiple contact points between two bodies’ surfaces. If multiple bodies share the same occurrence in the assembly hierarchy, they are given a pair-wise \code{Hierarchical} edge.

Once the graphs are constructed, graphs with less than three nodes or less than two edges are discarded, resulting in 4210 valid assembly graphs with 85,089 nodes and 118,668 edges. The encoded semantic embedding, the physical properties, and the geometric information are concatenated into a node feature matrix for each graph. Similarly, the one-hot encoded connection types are concatenated into an edge feature matrix. The node and edge feature matrices and the connectivity information represented in coordinate (COO) format are organized into PyTorch Geometric \cite{Fey2019} graph objects and are later fed into the GNN model.

\subsection{Learning Framework}\label{sec:learning_framework}
Given a set of graphs $\mathcal{G}=\left\{G_k\right\}_{k=1}^N$ where each graph $G_k = \left(V, E, \textbf{X}\right)$ 
consists of $|V|$ 
nodes, $|E|$ edges ($E \subseteq V \times V$), initial node features $\textbf{X}\in \mathbb{R}^{|V| \times d_x}$ and 
edge attributes $e_{uv}$ for $(u, v) \in E$, and the corresponding node labels 
$\left\{y_{1}^1, . . . ,y_{|V|}^1, . . . ,y_{1}^N, . . . , y_{|V|}^N\right\}$, the task of supervised node 
classification is to learn a representation $h_v, \forall v \in V$ such that the node labels can be predicted using 
the representations. GNNs use a neighborhood aggregation approach, where representation of node $v$ is iteratively 
updated by aggregating representations of neighboring nodes and edges. After $k$ iterations of aggregation, the 
representation captures the information within its $k$-hop neighborhood \cite{gilmer_2017_icml, hassani2020contrastive}. 
Formally, the $k$-th layer of a GNN is defines as:
\begin{equation}
h_v^{(k)} = f_\theta^{(k)}\left(h_v^{(k-1)}, g_\phi^{(k)}\left(\left\{h_v^{(k-1)}, h_u^{(k-1)}, e_{uv} : u \in N(v) \right\}\right.\right)
\end{equation} 
where $h_v^{(k)}$ is the representation of node $v$ at the $k$-th layer and $N(v)$ denote neighbors of $v$. $f_\theta(.)$ and $g_\phi(.)$ denote  parametric combination and aggregation functions. The choice of these functions can dramatically affect the expressiveness power of a GNN and different instantiations of them produce variants of GNNs such as GraphSAGE 
\cite{hamilton_2017_nips}, Graph Convolution Network (GCN) \cite{kipf_2017_iclr}, Graph Attention Network (GAT) 
\cite{velickovic_2018_iclr}, and Graph Isomorphism Network (GIN) \cite{xu_2019_iclr}.

The learning framework consists of a GNN encoder followed by a Multi-Layer Perceptron (MLP) network acting as a classifier head (Figure \ref{fig:framework}). 
% Specifically, we pass the graph connectivity information encoded in adjacency lists together with the node and edge features encoded in concatenated vectors into the GNN encoder. After the message passing step of each layer, we obtain the node and edge embeddings and perform non-linear activation using the Leaky Rectified Linear Unit \cite{Maas13rectifiernonlinearities}. For generating the final node embeddings to perform node classifications on the input graph, we follow a similar procedure as the Jumping Knowledge Networks \cite{xu2018representation} by aggregating the node embeddings of individual GNN layers from the learning framework through max-pooling.
The GNN encoder consists of a message-passing block in each layer followed by a Leaky Rectified Linear Unit \cite{Maas13rectifiernonlinearities} as the non-linearity. The final node embeddings are computed similar to Jumping Knowledge Networks \cite{xu2018representation} in which node embeddings computed in all the GNN layers are aggregated by summation. The final node embeddings are then passed through the MLP with batch normalization layers \cite{ioffe2015batch},  Parametric Rectified Linear Unit (PReLU \cite{he2015delving}) for generalized and learnable non-linearity activation, and a softmax activation for predicting class distributions. The weighted cross-entropy loss is adopted as the training objective to prevent the neural network from overlooking rare classes. The weights for each class are initialized as inversely proportional to the ground-truth class frequencies. The type of GNN layers is considered as a hyperparameter (see \ref{sec:general-setup}).

% We use a two-phase learning framework. In phase 1, we pre-train the MVCNN using the material categories. We then remove the classifier head of the MVCNN,
% freeze its weights, and use the generated visual encodings as part of initial node features.  In phase 2, we train the GNN using the same material categories. 
% The GNN architecture consists of a few layers of GIN and Batch normalization layers. The node encodings of all the layers are aggregated to represent the final
% node encodings (similar to Jumping Knowledge Networks \cite{xu2018representation}). In both MVCNN and GNN models, we use weighted cross entropy loss to address
% the class imbalance.

%%%%%%%%%%%%%%%%%%%%%%%%%%%%%%%%%%%%%%%%%%%%%%%%%%%%%%%%%%%%%%%%%%%%%%

\section{EXPERIMENTS}
\label{sec:experiments}
We evaluate and analyze the effectiveness of our framework with the Fusion 360 Gallery dataset. We designed three experiments through a variation of our base framework to address engineers' diverse expertise, needs, and goals. We first introduce a general setup, including the results of hyperparameter tuning. Then we discuss the feature importance in material prediction through ablation experiments. Finally, we provide the detailed motivation, description, results, and discussion for each experiment. 
We then address the following questions: (1) Which features have the largest influence on material prediction? (2) How well can our method predict the ground-truth material when observing the top-1, top-2, and top-3 predictions? (3) How does our method perform when the material of some of the bodies in the assembly is known? (4) Can the user restrict the search space by guiding the network towards a specific material category for each body?

\subsection{General Setup}
\label{sec:general-setup}
In preparation for the training process, we split the constructed assembly graphs into a train set for model training, a validation set for parameter tuning, and a test set for performance evaluation based on the proportions of 56\%, 24\%, 20\% respectively through a random sampling process. The assembly graphs allocated to the test set are pre-defined by the Fusion 360 Gallery dataset and remain fixed across the dataset-splitting process of different experiments. We run each experiment for 10 iterations to account for variations and evaluate the performance based on the mean and standard deviation of the results. To evaluate the multi-class classification task of material prediction, considering the highly-skewed distribution of material labels (as shown in Appendix \ref{appendix:dataset-statistics}), we report the \textit{micro} $F_1$ score calculated by weighting each prediction instance equally. 
% To address the possible impact of class imbalance, we also provide the \textit{weighted} perspective of the $F_1$ score, which takes into account the discrepancies between class sizes.

To find the best set of hyper-parameters for the GNN, we ran a grid search on the number of GNN layers $\left\{1, 2, 3, 4, 5, 6, 7, 8\right\}$, the size of hidden dimensions $\left\{64, 128, 256, 512\right\}$, and the type of GNN layer $\left\{SAGE, GAT, GIN, GCN\right\}$ (as introduced in Section \ref{sec:GNN}). After running each experiment configured with a combination of hyper-parameters for 10 iterations, we observed the best average performance and stability using 7 layers of GraphSAGE GNN with 256 hidden dimensions, where the formulation and advantage for the GraphSAGE GNN are described in Appendix \ref{sec:gnn_formulation}. Therefore, we adopted this configuration for the following experiments unless stated otherwise.

\subsection{Feature Importance}
\paragraph{Description}
To evaluate the importance of the features introduced to the graph construction and learning process, we perform an ablation study by systematically analyzing the variation of prediction performance by dropping sets of node and edge features. Specifically, we investigate the importance of node features by excluding only one node feature at a time during graph generation and learning. Subsequently, we re-run the node feature ablation experiments under the scenarios of no edge ablation and \code{Hierarchical} edge ablation to investigate the impact of edge features on the overall performance. Finally, we provide a set of baselines by including all node features.

\begin{table}[b] % \vspace{-0.7\baselineskip}
\setlength{\tabcolsep}{3pt}
\caption{\uppercase{\footnotesize Feature ablation results, micro-$F_1$ score}}
\label{table:feature-importance}
\vskip 0.05in
\begin{center}
\begin{footnotesize}
\begin{sc}
\begin{tabular}{l|cc}
\toprule
\multicolumn{1}{c}{Node Ablation} & \multicolumn{2}{c}{Edge Ablation}\\
\midrule
Node & None & Hierarchical\\
\midrule
Body Name & 0.384 $\pm$ 0.02 & 0.416 $\pm$ 0.03 \\
Occurrence Name & 0.399 $\pm$ 0.01 & 0.423 $\pm$ 0.01 \\
Semantic Names & 0.317 $\pm$ 0.05 & 0.373 $\pm$ 0.07 \\
Body Physical Properties & 0.413 $\pm$ 0.02 & 0.425 $\pm$ 0.01\\
Occurrence Physical Properties & 0.415 $\pm$ 0.02 & 0.392 $\pm$ 0.06\\
Body Geometry & 0.394 $\pm$ 0.02 & 0.420 $\pm$ 0.00 \\
Global Features & 0.393 $\pm$ 0.04 & 0.429 $\pm$ 0.01  \\
\midrule
None & 0.404 $\pm$ 0.02 & 0.425 $\pm$ 0.01 \\
% All & ? & ? \\
\bottomrule
\end{tabular}
\end{sc}
\end{footnotesize}
\end{center}
% \vskip -0.1in
\end{table}

\paragraph{Results and Discussion}
Shown in Table \ref{table:feature-importance} is the feature ablation results, where \code{SEMANTIC NAMES} indicates the body and occurrence names of the node, \code{NONE} and \code{ALL} indicate no feature and all features being ablated, respectively. 

Focusing on the scenario in which \textit{no ablation of edges is performed}, we observe that the most important node feature is the \code{semantic names} of the body since the exclusion of it significantly reduced the averaged performance (from $0.404$ to $0.317$) and increased the standard deviation (from $0.02$ to $0.05$). Dissecting the semantic names feature, we observe that both \code{body name} and \code{occurrence name} impact the accuracy when ablated, where the \code{body name}'s importance surpasses that of the \code{occurrence name}. Furthermore, excluding \code{physical properties} reflects an increase in averaged performance, which indicates that they are hindering the learning process. This phenomenon is unexpected since the physical properties of an assembly body should be correlated with its material choice. Therefore, we choose to re-run the ablation experiments without considering the hierarchical edges to investigate this adversarial effect. 

When \textit{ablating hierarchical edges}, we observe that the \code{semantic names} node feature shows the most significance, which is in accordance with the previous result. Consequently, we suggest that the user-defined semantic names of bodies in CAD models are the single most useful feature for material prediction. Further research should be performed to investigate more effective ways to extract and encode this feature. In addition, the physical properties of the body and occurrence demonstrate significance, which is contrary to the result when no edge ablation is performed. One possible explanation would be that the user-defined hierarchical edges are connecting bodies that may not be correlated with each other in terms of structural and physical properties. Accordingly, the introduction of hierarchical edges may misguide the neural network to learn representations of physical properties based on incorrect structural context. Therefore, we drop the hierarchical edges during subsequent experiments to mitigate this effect and minimize the computation overhead.

\subsection{Fully Algorithm-guided Prediction}

% Figure comparing: SAGE, MLP, Linear, MVCNN on Top 1, 3, 5.

\paragraph{Description}
A long-term goal of design automation is to fully automate the material selection process without user input. This experiment evaluates how our network performs in a fully algorithm-guided scenario, where the ground-truth labels of material features for all nodes are left out during training and are re-introduced only during the validation and testing phases. Therefore, the GNN learns the node representations of the body nodes without access to ground-truth materials, aggregates the single-node representations obtained through message passing directly into a joint set of representations, and performs material prediction on all graph nodes at the same time. During the evaluation, we treat the unseen assembly graphs as newly produced user designs and allow the trained model to output the top-1, top-2, and top-3 material class labels for all bodies based on the predicted probability. We denote this approach as the \textbf{fully algorithm-guided prediction} since the entire learning process is done by the learning framework independently without any supporting input from the user. The predictions may provide a set of suggestions for the user to select for further analysis or use. 

\begin{figure}[th]
\begin{center}
\includegraphics[width=8.5cm]{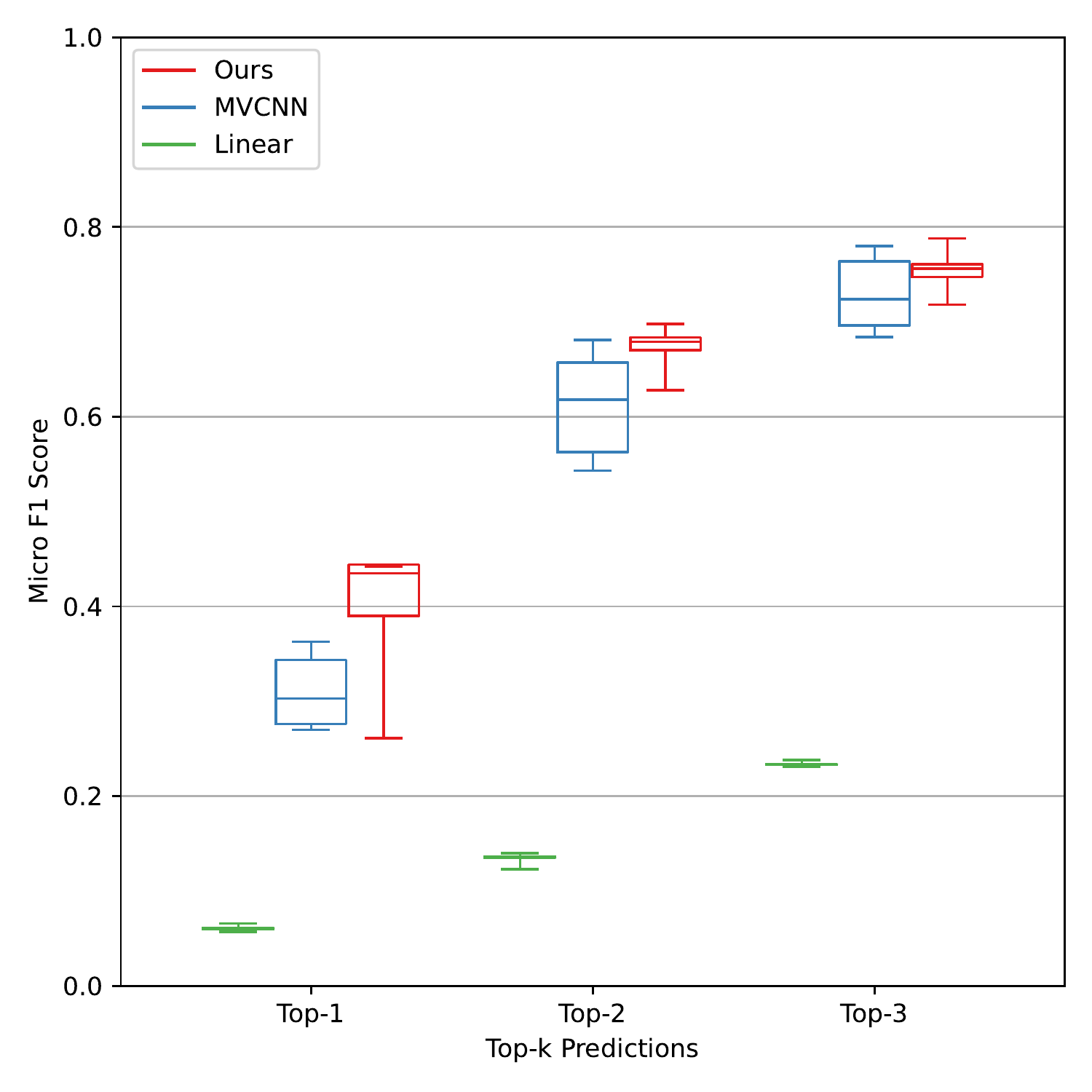}
\end{center}
\caption{\uppercase{\footnotesize Fully algorithm-guided prediction results, compared with several baselines.}}
\label{fig:fully-algorithm} 
\end{figure}

\paragraph{Results and Discussion}

As shown in Figure \ref{fig:fully-algorithm}, we compare the material prediction performance of our proposed method with the baselines of MVCNN (which performs prediction based only on the geometry of bodies) and a linear regression model. When applying this work, the top-k method offers the flexibility to use a human-in-the-loop approach. In practice, this would streamline the decision process while leaving some creative decision-making to the experienced designer. 
Our method achieves a mean micro-$F_1$ score of $0.417\pm0.04$, $0.677\pm0.01$, $0.754\pm0.01$ for top-1, top-2, and top-3 predictions respectively, and outperforms the baselines in all top-k situations. This supports our hypothesis that material selection depends on factors other than the geometric features of the design. Designers choose the materials of one body in an assembly also based on the context of the rest of the assembly.  

As more labels are considered in the prediction, the MVCNN approach appears to converge with our method, which is expected as two main classes dominate our dataset. Nonetheless, the application limits the MVCNN as it only learns from the geometry of prior designs, not from other multi-modal design data. 
In the remaining two experiments, we consider more input from the user, which can be easily introduced into the model.

\subsection{Partial Algorithm-guided Prediction}

\paragraph{Description}
Designers may have access to material labels for part of their assembly or design only one body in an existing assembly. To simulate this, this experiment introduces ground-truth material labels as node features into a portion of assembly graph nodes (\textit{context} nodes), and the goal is to learn the labels for the rest (\textit{target} nodes). During the message-passing process, the \textit{context} nodes pass their embeddings through the edges to the \textit{target} nodes, thereby updating their embeddings. This allows the GNN to perform material prediction on the \textit{target} nodes by jointly learning from the topology of the graph and the ground-truth labels contained within the \textit{context} nodes. During the evaluation, we limit the number of \textit{context} nodes to a set of ratios and analyze the change in prediction performance jointly with a different number of GNN layers. We denote this as \textbf{partial algorithm-guided prediction} since the ground-truth labels provided by the user can influence, to a certain degree, the learning process and bias the final predictions. While the fully algorithm-guided prediction has the ground truth information of all nodes removed during training, the partial algorithm-guided prediction allows for a portion of context nodes to contain the ground truth labels during training but does not consider them during validation.
% Since a portion of the ground-truth labels is assumed to be present in the assembly graph, it simulates the application where the designers have preliminary knowledge of some bodies in the design and use the model to provide suggestions for a subset of specific bodies, biasing the suggestions with the context of neighboring bodies.

\begin{table}[h]
\setlength{\tabcolsep}{3pt}
\begin{footnotesize}
\caption{\uppercase{\footnotesize Partial-algorithm Prediction Results \newline (micro-$f_1$score).}}
\label{table:partial-experiment}
\vskip 0.05in
\begin{center}
\begin{tabular}{cccccccccc}
\addlinespace[-\aboverulesep] 
\cmidrule[\heavyrulewidth]{3-10}
\multicolumn{1}{l}{} &  & \multicolumn{8}{c}{\textbf{Number of Layers}} \\
\cline{3-10}
\multicolumn{1}{l}{} & \multirow{-2}{*}{\textbf{}} & \textbf{1} & \textbf{2} & \textbf{3} & \textbf{4} & \textbf{5} & \textbf{6} & \textbf{7} & \textbf{8} \\ \midrule
\multicolumn{1}{c|}{} & \textbf{0.1} & \cellcolor[HTML]{DFE596}0.402 & \cellcolor[HTML]{E6E897}0.397 & \cellcolor[HTML]{E6E797}0.398 & \cellcolor[HTML]{DDE595}0.403 & \cellcolor[HTML]{A1D289}0.442 & \cellcolor[HTML]{BCDA8E}0.425 & \cellcolor[HTML]{DCE495}0.404 & \cellcolor[HTML]{E3E696}0.400 \\
\multicolumn{1}{c|}{} & \textbf{0.2} & \cellcolor[HTML]{F7ED9B}0.386 & \cellcolor[HTML]{C3DC90}0.421 & \cellcolor[HTML]{CCDF92}0.414 & \cellcolor[HTML]{ADD68B}0.435 & \cellcolor[HTML]{C9DE91}0.416 & \cellcolor[HTML]{96CE86}0.450 & \cellcolor[HTML]{F3EC9A}0.389 & \cellcolor[HTML]{FAEE9B}0.385 \\
\multicolumn{1}{c|}{} & \textbf{0.3} & \cellcolor[HTML]{FFEF9C}0.381 & \cellcolor[HTML]{EDEA99}0.393 & \cellcolor[HTML]{CBDF91}0.415 & \cellcolor[HTML]{86C983}0.460 & \cellcolor[HTML]{6DC27E}0.476 & \cellcolor[HTML]{78C580}0.469 & \cellcolor[HTML]{7DC681}0.466 & \cellcolor[HTML]{B2D78C}0.431 \\
\multicolumn{1}{c|}{} & \textbf{0.4} & \cellcolor[HTML]{E0E696}0.402 & \cellcolor[HTML]{E4E797}0.399 & \cellcolor[HTML]{ACD58B}0.435 & \cellcolor[HTML]{83C882}0.462 & \cellcolor[HTML]{6FC27E}0.475 & \cellcolor[HTML]{6CC17D}0.477 & \cellcolor[HTML]{71C37E}0.474 & \cellcolor[HTML]{88CA83}0.459 \\
\multicolumn{1}{c|}{\multirow{-1.6}{*}{\textbf{\multirot{Context Nodes \%}}}}  & \textbf{0.5} & \cellcolor[HTML]{EEEA99}0.393 & \cellcolor[HTML]{D8E394}0.407 & \cellcolor[HTML]{AAD58B}0.436 & \cellcolor[HTML]{93CD86}0.452 & \cellcolor[HTML]{64BF7C}0.482 & \cellcolor[HTML]{80C781}0.464 & \cellcolor[HTML]{63BE7B}0.482 & \cellcolor[HTML]{6EC27E}0.475 \\
\bottomrule
\end{tabular}
\end{center}
\end{footnotesize}
\end{table}

\paragraph{Results and Discussion}

Table \ref{table:partial-experiment} summarizes the results of the experiment, showing the averaged micro-$F_1$ score for the top-1 prediction. We observe that the performance increases as we increase the ratio of augmented nodes and the number of GNN layers. The positive correlation between the ratio and the performance indicates that the model can constructively learn from augmented nodes' topology and features during training and generalize at inference time. Similarly, the positive correlation between the number of GNN layers and the performance implies that the model learns better representations by considering the messages passed from distant nodes and edges in each graph. We observe a slight decrease in performance when layers exceed 7 due to over-parametrization.
% , where the message passing process reaches the periphery of the graph and ``bounces back", thereby retrieving redundant information and causing confoundment. 
In summary, the partial algorithm-guided prediction achieves the best performance in top-1 prediction of  $0.482\pm0.02$  when considering $50\%$ augmented nodes and $5$ GNN layers, indicating the method's effectiveness in making predictions based on partial inputs from the user.

\subsection{User-guided Prediction}
\paragraph{Description}
The designer might want to restrict the search space to a specific material category for each body, such as ``metal", ``ceramic", ``aluminum alloy", or ``thermoplastics". With a \textbf{user-guided prediction} approach, the user provides the network with ground-truth material categories for all bodies, thus guiding the network towards groups of materials based on their expertise, ideas, and design goals. The motivation of this experiment, together with the prior partial algorithm-guided experiment, is to avoid the limitation of innovation by allowing the user to input their design information into the learning framework, thereby leading the design process. Due to the lack of newly generated data by designers, we simulate the user inputs via masking and involving ground truth information in the graph generation process. The configuration of this experiment follows from the format of the fully algorithm-guided prediction. However, to aid the learning process, we introduce the hierarchical material category label of the ground-truth material as a feature of each node.

The material categories are borrowed from the material library in Autodesk Fusion 360 and consist of up to three tiers in the hierarchy, where a deeper depth indicates a more detailed classification. For example, a body made of mild steel has a Tier 1 material category ``Metal", Tier 2 category ``Ferrous", and Tier 3 category ``Carbon Steel". During the evaluation of this experiment, we limit the hierarchy of ground-truth material categories to a set of depths and analyze the change in performance based on the top-1, top-2, and top-3 prediction accuracy.

% With this \textbf{user-guided prediction} approach, the user provides the learning framework with the ground-truth material categories for all assembly bodies, which will heavily influence the result of predicted material labels. Since we are assuming that the input ground-truth material categories are correlated with the predicted material labels, we consider this approach suitable for the designers who would like to take control of the learning process and fully guide the network in material prediction based on their expertise, concept, or goal of design.

%%%%%%%%%%%%%%%% begin table %%%%%%%%%%%%%%%%%%%
\begin{table}[t] % \vspace{-0.7\baselineskip}
\setlength{\tabcolsep}{2.5pt}
\caption{\uppercase{\footnotesize User-guided Prediction Results.}}
\label{table:user-guided-results}
\vskip 0.05in
\begin{center}
\begin{footnotesize}
\begin{sc}
\begin{tabular}{cccccc}
\toprule

\textbf{Top-k} &\multicolumn{3}{c}{\textbf{Material Class}} & \multicolumn{2}{c}{\textbf{F$_1$ Score}}\\
\cmidrule{2-6}
~ & Tier 1 & Tier 2 & Tier 3 & Micro-($F_m$) & Weighted $(F_w)$ \\

\midrule
\multirow{4}{*}{\textbf{1}} 
& \ding{55} & \ding{55} & \ding{55} & 0.417 $\pm$ 0.04 & 0.392 $\pm$ 0.03\\
& \checkmark & \ding{55} & \ding{55} & 0.546 $\pm$ 0.01 & 0.527 $\pm$ 0.01\\
& \checkmark & \checkmark & \ding{55} & 0.736 $\pm$ 0.03 & 0.746 $\pm$ 0.03\\
& \checkmark & \checkmark & \checkmark & 0.731 $\pm$ 0.04 & 0.757 $\pm$ 0.03\\
\midrule
\multirow{4}{*}{\textbf{2}} 
& \ding{55} & \ding{55} & \ding{55} & 0.677 $\pm$ 0.01 & 0.630 $\pm$ 0.01\\
& \checkmark & \ding{55} & \ding{55} & 0.684 $\pm$ 0.10 & 0.677 $\pm$ 0.07\\
& \checkmark & \checkmark & \ding{55} & 0.841 $\pm$ 0.11 & 0.851 $\pm$ 0.09\\
& \checkmark & \checkmark & \checkmark & 0.897 $\pm$ 0.01 & 0.903 $\pm$ 0.01\\
\midrule

\multirow{4}{*}{\textbf{3}} 
& \ding{55} & \ding{55} & \ding{55} & 0.754 $\pm$ 0.01 & 0.704 $\pm$ 0.01\\
& \checkmark & \ding{55} & \ding{55} & 0.781 $\pm$ 0.12 & 0.763 $\pm$ 0.10\\
& \checkmark & \checkmark & \ding{55} & 0.889 $\pm$ 0.12 & 0.891 $\pm$ 0.12\\
& \checkmark & \checkmark & \checkmark & 0.953 $\pm$ 0.01 & 0.954 $\pm$ 0.01\\

\bottomrule
\end{tabular}
\end{sc}
\end{footnotesize}
\end{center}
% \vskip -0.1in
\end{table}
%%%%%%%%%%%%%%%% end table %%%%%%%%%%%%%%%%%%%

\paragraph{Results and Discussion}
Summarized in Table \ref{table:user-guided-results} are the user-guided prediction results. We observe the general trend of increasing performance with hierarchical material category input, which validates our hypothesis that the user-input material class information can effectively guide the model in making predictions. One phenomenon to notice is the convergence of performance when increasing the hierarchy of material class information with the number of prediction selections fixed. 
% For example, when including the tier-3 material class information during the top-1 prediction, the $F_m$ score dropped slightly from $0.7358$ to $0.7318$. 
This indicates that the model can efficiently adapt to the user-input information, even if the information provided is minimal. Therefore, in practice, our framework can provide users with an accurate and compendious set of material selections guided by a limited amount of user-input material class information.

%%%%%%%%%%%%%%%%%%%%%%%%%%%%%%%%%%%%%%%%%%%%%%%%%%%%%%%%%%%%%%%%%%%%%%
\section{DISCUSSION}

\subsection{Limitations and Improvement}
\paragraph{Class Imbalance}
As shown in Table~\ref{table:material-count} of Appendix~\ref{appendix:dataset-statistics}, the distribution of the ground-truth materials is highly skewed. Specifically, the default material (Steel) significantly dominates over the less common ones (i.e., Brass). The lack of representative examples from the minority class and the overwhelming amount of default cases less relevant to the task will misguide the model to make predictions biased towards the majority class. Despite the efforts made to mitigate the impact of imbalanced material labels, such as down-sampling the data, grouping rare labels, and adopting weighted loss calculation, the issue persists.

We seek to collect additional data with a more varied distribution of features and ground-truth materials for improvement. This could be achieved either by expanding the currently adopted Fusion 360 Gallery Assembly Dataset via inviting more designers to contribute or by using publicly available datasets such as ABC \cite{Koch_2019_CVPR}. Furthermore, data augmentation through manipulating existing graphs, such as creating sub-graphs or synthesized graphs from past designs, may be a viable direction for improving the distribution of material classes. Existing tools designed to support automated reasoning, such as Open-NARS \cite{10.5555/1565455.1565462}, could also be applied to the graph representation learning process to support the inference and imputation of rare material classes. Furthermore, we are working on re-categorizing the hierarchical categories of material labels to produce a more reasonable distribution while preserving the soundness of the designs.

% focal loss, increasing the weight factor of the loss

\paragraph{Functional and Behavioral Features}
Our current work mainly focuses on the \textit{structural} aspect of material selection, such as the geometric and physical properties of bodies, as well as the assembly relationships that correlate them. While structural analysis has proven to be successful in the derivation and tabulation of material performance indices for standard mechanical design cases \cite{ashby_materials_2011}, the \textit{functional} and \textit{behavioral} aspects of the design are also highly influential to material selection \cite{article, arnold2012materials}. Therefore, one promising improvement would be to incorporate the functional information (i.e., the purpose) and the behavioral information (i.e., the attributes) of bodies as node features to narrow down the search space. While the TechNet embeddings of the semantic names might be implicitly representing some functional and behavioral aspects, introducing new features allows for a refined feature encoding process that can more effectively reflect the multifaceted information relevant to material selection in design automation, thereby enhancing the overall performance and robustness of the proposed framework when dealing with multi-modal real-world inputs.

% https://docs.google.com/presentation/d/1IPDRH_gG_DsZfTDBdM0IMG2CEwnWYCNM/edit#slide=id.p3

% \paragraph{Overfitting}
% [learning framework improvements] Fix the overfitting?

\subsection{Future Plans}\label{sec:future_plans}

% [OpenNARS can work with missing data, unbalanced data. NN requires a lot of data, and high-quality data. Incorporating the reasoning system might provide a more balanced approach. (deduction, abduction, induction reasoning).]

\paragraph{Regression for Material Properties}
One of the limitations of the proposed framework is the dependency on a specific material library found in Autodesk Fusion 360. This dependency comes from framing the problem as a node classification task, where the number of possible classes is equal to the number of materials in the library, which for this dataset is 576. The categorical classification is challenging due to the number of possible classes. It also limits the application of the trained model to only materials found in the original dataset, which is incomplete and might not suit the needs of designers in different industries. A possible improvement might be to frame the problem as a regression task and develop a model to predict relevant physical and mechanical properties involved in the selection of materials, building on prior work that leveraged neural networks to predict the density and Young's modulus of materials given their chemical compositions \cite{merayo2020prediction}. By mapping relevant material properties of the material library of the training data onto an Ashby chart, clustering of the material properties on the chart would enable a more flexible material selection method.

\paragraph{Graph Predictions} 
Given the proposed framework's ability to capture structural characteristics of CAD models via graph construction and learning relational information of design features through graph representation learning, we envision the potential expansion of our work in producing diverse predictions for design automation that is not limited to material selection. Specifically, we plan to incorporate graph node prediction, graph edge prediction, and global context prediction experiments through a variation of our current graph representation learning process. For \textbf{graph node prediction task}, we aim to automate the user-design process of assembly bodies by providing suggestions regarding the structural, functional, and behavioral features through a joint representation learning of node, edge, and global graph features. The current material prediction task falls into node prediction and may serve as a relevant foundation. For \textbf{graph edge prediction task}, we aim to support the user-design process of organizing and correlating assembly bodies by providing insights into their hierarchical and relational information through representation learning based on their properties and that of the assembly. For \textbf{global context prediction}, we aim to provide users with an informative overview of the entire assembly design through representation learning tailored to the global graph features shared across individual bodies. The overview of the assembly design we provide may adapt to the introduction of newly designed bodies, thereby serving as a flexible auxiliary unit to address the needs of different users.

%%%%%%%%%%%%%%%%%%%%%%%%%%%%%%%%%%%%%%%%%%%%%%%%%%%%%%%%%%%%%%%%%%%%%%
    \section{CONCLUSION}
    
    In this paper, we propose a unified framework that contributes to design automation by predicting the material of bodies in assemblies through graph representation learning, given user design knowledge input. Furthermore, we develop a systematic workflow for the feature extraction, encoding, and assembly graph construction of CAD models from the Fusion 360 Gallery Assembly Dataset. 
    
    During the experimental evaluation, we present three experiments through a variation of our base framework tailored to the diverse needs of designers. For the fully algorithm-guided experiment, in which predictions of all body materials are performed simultaneously without any ground-truth input, our model achieved a micro-$F_1$ score of 0.417, 0.677, 0.754 for top-1, top-2, and top-3 predictions, respectively, surpassing those of the MVCNN baseline. For the partial algorithm-guided experiment, in which predictions of target nodes are based on a joint representation learning of graph topology and the ground truths of neighbor nodes, our model achieved 0.482 top-1 micro-$F_1$. For the user-guided experiment, in which the user's input of ground-truth material categories take part in the learning process, our model achieved an averaged micro-$F_1$ score of 0.731 for top-1 prediction and up to 0.953 for top-3 predictions.
    
    While our results demonstrate the feasibility of leveraging graph representation learning for feature predictions on graphically represented CAD models, the configuration of the three experiments shows promise in supporting human-in-the-loop design automation applications. Specifically, we believe the proposed framework's capacity to accommodate large-scale databases and flexibility in incorporating the designer's knowledge can be used to create a recommendation system for users by learning best practices from existing designs. Furthermore, the framework may serve as a baseline for future works leveraging graph neural networks for design automation.
    
    We view material selection as one step towards a fully automated data-driven design tool capable of synthesizing assemblies from design requirements. We hope to support further research to bridge the gap between the understanding of human designers and that of intelligent design agents.

%%%%%%%%%%%%%%%%%%%%%%%%%%%%%%%%%%%%%%%%%%%%%%%%%%%%%%%%%%%%%%%%%%%%%%

% Here's where you specify the bibliography style file.
% The full file name for the bibliography style file 
% used for an ASME paper is asmems4.bst.
\bibliographystyle{asmems4}

%%%%%%%%%%%%%%%%%%%%%%%%%%%%%%%%%%%%%%%%%%%%%%%%%%%%%%%%%%%%%%%%%%%%%%
\begin{acknowledgment}
This research was supported by the U.S. NASA MUREP Institutional Research Opportunity (MIRO) program (Award Number 80NSSC19M0200), donation from the Autodesk Inc, and partially carried out at the Jet Propulsion Laboratory, California Institute of Technology, under a contract with the National Aeronautics and Space Administration (80NM0018D0004).
\end{acknowledgment}

%%%%%%%%%%%%%%%%%%%%%%%%%%%%%%%%%%%%%%%%%%%%%%%%%%%%%%%%%%%%%%%%%%%%%%

% Here's where you specify the bibliography database file.
% The full file name of the bibliography database for this
% article is asme2e.bib. The name for your database is up
% to you.
\bibliography{asme2e}

\begin{thebibliography}{10}

\bibitem{arnold2012materials}
Arnold, S.~M., Cebon, D., and Ashby, M., 2012.
\newblock ``Materials selection for aerospace systems''.

\bibitem{mouritz2012introduction}
Mouritz, A., 2012.
\newblock {\em Introduction to Aerospace Materials}.
\newblock Woodhead Publishing in Materials. Elsevier Science.

\bibitem{10.31399/asm.hb.v11.a0003501}
Miller, B.~A., 2002.
\newblock ``{Materials Selection for Failure Prevention}''.
\newblock In {\em {Failure Analysis and Prevention}}. ASM International, 01.

\bibitem{Sarica2020}
Sarica, S., Luo, J., and Wood, K.~L., 2020.
\newblock ``{TechNet: Technology semantic network based on patent data}''.
\newblock {\em Expert Systems with Applications, \textbf{ 142}}.

\bibitem{ferrero2022classifying}
Ferrero, V., DuPont, B., Hassani, K., and Grandi, D., 2022.
\newblock ``Classifying component function in product assemblies with graph
  neural networks''.
\newblock {\em Journal of Mechanical Design, \textbf{ 144}}(2).

\bibitem{siddharth2022engineering}
Siddharth, L., Blessing, L., Wood, K.~L., and Luo, J., 2022.
\newblock ``Engineering knowledge graph from patent database''.
\newblock {\em Journal of Computing and Information Science in Engineering,
  \textbf{ 22}}(2).

\bibitem{sargent2016materials}
Sargent, P., 2016.
\newblock {\em Materials information for CAD/CAM}.
\newblock Elsevier.

\bibitem{willis2020fusion}
Willis, K. D.~D., Pu, Y., Luo, J., Chu, H., Du, T., Lambourne, J.~G.,
  Solar-Lezama, A., and Matusik, W., 2021.
\newblock ``Fusion 360 gallery: A dataset and environment for programmatic cad
  construction from human design sequences''.
\newblock {\em ACM Transactions on Graphics (TOG), \textbf{ 40}}(4).

\bibitem{Koch_2019_CVPR}
Koch, S., Matveev, A., Jiang, Z., Williams, F., Artemov, A., Burnaev, E.,
  Alexa, M., Zorin, D., and Panozzo, D., 2019.
\newblock ``Abc: A big cad model dataset for geometric deep learning''.
\newblock In The IEEE Conference on Computer Vision and Pattern Recognition
  (CVPR).

\bibitem{sangpil2020large}
Kim, S., Chi, H.-g., Hu, X., Huang, Q., and Ramani, K., 2020.
\newblock ``A large-scale annotated mechanical components benchmark for
  classification and retrieval tasks with deep neural networks''.
\newblock In Proceedings of 16th European Conference on Computer Vision (ECCV).

\bibitem{willis2021joinable}
Willis, K. D.~D., Jayaraman, P.~K., Chu, H., Tian, Y., Li, Y., Grandi, D.,
  Sanghi, A., Tran, L., Lambourne, J.~G., Solar-Lezama, A., and Matusik, W.,
  2021.
\newblock Joinable: Learning bottom-up assembly of parametric cad joints.

\bibitem{jones2021sb}
Jones, B., Hildreth, D., Chen, D., Baran, I., Kim, V., and Schulz, A., 2021.
\newblock ``Sb-gcn: Structured brep graph convolutional network for automatic
  mating of cad assemblies''.
\newblock {\em arXiv preprint arXiv:2105.12238}.

\bibitem{lambourne2021brepnet}
Lambourne, J.~G., Willis, K.~D., Jayaraman, P.~K., Sanghi, A., Meltzer, P., and
  Shayani, H., 2021.
\newblock ``Brepnet: A topological message passing system for solid models''.
\newblock In Proceedings of the IEEE/CVF Conference on Computer Vision and
  Pattern Recognition (CVPR), pp.~12773--12782.

\bibitem{jayaraman2020uv}
Jayaraman, P.~K., Sanghi, A., Lambourne, J., Davies, T., Shayani, H., and
  Morris, N., 2020.
\newblock ``Uv-net: Learning from curve-networks and solids''.
\newblock {\em arXiv preprint arXiv:2006.10211}.

\bibitem{doi:10.1179/mst.1989.5.6.517}
Ashby, M.~F., 1989.
\newblock ``Materials selection in conceptual design''.
\newblock {\em Materials Science and Technology, \textbf{ 5}}(6), pp.~517--525.

\bibitem{ullah2020investigation}
Ullah, N., Riaz, A.~A., and Shah, S.~A., 2020.
\newblock ``Investigation on material selection for the columns of universal
  testing machine (utm) using granta’s design ces edupack''.
\newblock {\em Technical Journal, \textbf{ 25}}(02), pp.~52--60.

\bibitem{ALBINANA2012433}
Albiñana, J., and Vila, C., 2012.
\newblock ``A framework for concurrent material and process selection during
  conceptual product design stages''.
\newblock {\em Materials \& Design, \textbf{ 41}}, pp.~433--446.

\bibitem{van_kesteren_2008}
van Kesteren, I., 2008.
\newblock {\em Materials, user-product interaction and other decisions}.

\bibitem{zarandi2011material}
Zarandi, M. H.~F., Mansour, S., Hosseinijou, S.~A., and Avazbeigi, M., 2011.
\newblock ``A material selection methodology and expert system for sustainable
  product design''.
\newblock {\em The International Journal of Advanced Manufacturing Technology,
  \textbf{ 57}}(9-12), pp.~885--903.

\bibitem{doi:10.1080/002075431000087265}
Abdullah, T.~A., Popplewell, K., and Page, C.~J., 2003.
\newblock ``A review of the support tools for the process of assembly method
  selection and assembly planning''.
\newblock {\em International Journal of Production Research, \textbf{ 41}}(11),
  pp.~2391--2410.

\bibitem{li_2015_iclr}
Li, Y., Tarlow, D., Brockschmidt, M., and Zemel, R., 2015.
\newblock ``Gated graph sequence neural networks''.
\newblock In International Conference on Learning Representations.

\bibitem{gilmer_2017_icml}
Gilmer, J., Schoenholz, S.~S., Riley, P.~F., Vinyals, O., and Dahl, G.~E.,
  2017.
\newblock ``Neural message passing for quantum chemistry''.
\newblock In International Conference on Machine Learning, pp.~1263--1272.

\bibitem{kipf_2017_iclr}
Kipf, T.~N., and Welling, M., 2017.
\newblock ``Semi-supervised classification with graph convolutional networks''.
\newblock In International Conference on Learning Representations.

\bibitem{velickovic_2018_iclr}
Veličković, P., Cucurull, G., Casanova, A., Romero, A., Liò, P., and Bengio,
  Y., 2018.
\newblock ``Graph attention networks''.
\newblock In International Conference on Learning Representations.

\bibitem{xu_2019_iclr}
Xu, K., Hu, W., Leskovec, J., and Jegelka, S., 2019.
\newblock ``How powerful are graph neural networks?''.
\newblock In International Conference on Learning Representations.

\bibitem{Khasahmadi2020Memory}
Khasahmadi, A.~H., Hassani, K., Moradi, P., Lee, L., and Morris, Q., 2020.
\newblock ``Memory-based graph networks''.
\newblock In International Conference on Learning Representations.

\bibitem{hassani_2019_iccv}
Hassani, K., and Haley, M., 2019.
\newblock ``Unsupervised multi-task feature learning on point clouds''.
\newblock In International Conference on Computer Vision, pp.~8160--8171.

\bibitem{wang_2018_iclr}
Wang, T., Zhou, Y., Fidler, S., and Ba, J., 2019.
\newblock ``Neural graph evolution: Automatic robot design''.
\newblock In International Conference on Learning Representations.

\bibitem{sanchez_2018_icml}
Sanchez-Gonzalez, A., Heess, N., Springenberg, J.~T., Merel, J., Riedmiller,
  M., Hadsell, R., and Battaglia, P., 2018.
\newblock ``Graph networks as learnable physics engines for inference and
  control''.
\newblock In International Conference on Machine Learning, pp.~4470--4479.

\bibitem{sanchez2020learning}
Sanchez-Gonzalez, A., Godwin, J., Pfaff, T., Ying, R., Leskovec, J., and
  Battaglia, P., 2020.
\newblock ``Learning to simulate complex physics with graph networks''.
\newblock In International Conference on Machine Learning, PMLR,
  pp.~8459--8468.

\bibitem{gilmer2017neural}
Gilmer, J., Schoenholz, S.~S., Riley, P.~F., Vinyals, O., and Dahl, G.~E.,
  2017.
\newblock Neural message passing for quantum chemistry.

\bibitem{guo2020semi}
Guo, K., and Buehler, M.~J., 2020.
\newblock ``A semi-supervised approach to architected materials design using
  graph neural networks''.
\newblock {\em Extreme Mechanics Letters, \textbf{ 41}}, p.~101029.

\bibitem{marcheggiani2017encoding}
Marcheggiani, D., and Titov, I., 2017.
\newblock Encoding sentences with graph convolutional networks for semantic
  role labeling.

\bibitem{ahmed2021graph}
Ahmed, F., Cui, Y., Fu, Y., and Chen, W., 2021.
\newblock A graph neural network approach for product relationship prediction.

\bibitem{doi:10.1080/00207543.2018.1443229}
Tao, F., Sui, F., Liu, A., Qi, Q., Zhang, M., Song, B., Guo, Z., Lu, S. C.-Y.,
  and Nee, A. Y.~C., 2019.
\newblock ``Digital twin-driven product design framework''.
\newblock {\em International Journal of Production Research, \textbf{ 57}}(12),
  pp.~3935--3953.

\bibitem{feng2020data}
Feng, Y., Zhao, Y., Zheng, H., Li, Z., and Tan, J., 2020.
\newblock ``Data-driven product design toward intelligent manufacturing: A
  review''.
\newblock {\em International Journal of Advanced Robotic Systems, \textbf{
  17}}(2), p.~1729881420911257.

\bibitem{funkhouser2004modeling}
Funkhouser, T., Kazhdan, M., Shilane, P., Min, P., Kiefer, W., Tal, A.,
  Rusinkiewicz, S., and Dobkin, D., 2004.
\newblock ``Modeling by example''.
\newblock {\em ACM transactions on graphics (TOG), \textbf{ 23}}(3),
  pp.~652--663.

\bibitem{Williams2019}
Williams, G., Meisel, N.~A., Simpson, T.~W., and McComb, C., 2019.
\newblock ``{Design repository effectiveness for 3D convolutional neural
  networks: Application to additive manufacturing}''.
\newblock {\em Journal of Mechanical Design, Transactions of the ASME, \textbf{
  141}}(11), pp.~1--12.

\bibitem{wang2021understanding}
Wang, Y., Grandi, D., Cui, D., Rao, V., and Goucher-Lambert, K., 2021.
\newblock ``Understanding professional designers’ knowledge organization
  behavior: A case study in product teardowns''.
\newblock In International Design Engineering Technical Conferences and
  Computers and Information in Engineering Conference, Vol.~85420, American
  Society of Mechanical Engineers, p.~V006T06A046.

\bibitem{chaudhuri2020learning}
Chaudhuri, S., Ritchie, D., Wu, J., Xu, K., and Zhang, H., 2020.
\newblock ``Learning generative models of 3d structures''.
\newblock Vol.~39, Wiley Online Library, pp.~643--666.

\bibitem{li2020learning}
Li, J., Niu, C., and Xu, K., 2020.
\newblock ``Learning part generation and assembly for structure-aware shape
  synthesis''.
\newblock Vol.~34, pp.~11362--11369.

\bibitem{mo2019structurenet}
Mo, K., Guerrero, P., Yi, L., Su, H., Wonka, P., Mitra, N., and Guibas, L.~J.,
  2019.
\newblock ``Structurenet: Hierarchical graph networks for 3d shape
  generation''.
\newblock {\em ACM Trans. Gr., \textbf{ 38}}(6).

\bibitem{slyadnev2020role}
Slyadnev, S., Malyshev, A., Voevodin, A., and Turlapov, V., 2020.
\newblock ``On the role of graph theory apparatus in a cad modeling kernel''.

\bibitem{HUANG20081073}
Huang, H., Lo, S., Zhi, G., and Yuen, R., 2008.
\newblock ``Graph theory-based approach for automatic recognition of cad
  data''.
\newblock {\em Engineering Applications of Artificial Intelligence, \textbf{
  21}}(7), pp.~1073--1079.

\bibitem{SUNIL2010686}
Sunil, V., Agarwal, R., and Pande, S., 2010.
\newblock ``An approach to recognize interacting features from b-rep cad models
  of prismatic machined parts using a hybrid (graph and rule based)
  technique''.
\newblock {\em Computers in Industry, \textbf{ 61}}(7), pp.~686--701.

\bibitem{su15mvcnn}
Su, H., Maji, S., Kalogerakis, E., and Learned{-}Miller, E.~G., 2015.
\newblock ``Multi-view convolutional neural networks for 3d shape
  recognition''.
\newblock In Proc. ICCV.

\bibitem{Fey2019}
Fey, M., and Lenssen, J.~E., 2019.
\newblock ``Fast graph representation learning with {PyTorch Geometric}''.
\newblock In ICLR Workshop on Representation Learning on Graphs and Manifolds.

\bibitem{hassani2020contrastive}
Hassani, K., and Khasahmadi, A.~H., 2020.
\newblock ``Contrastive multi-view representation learning on graphs''.
\newblock In International Conference on Machine Learning, pp.~4116--4126.

\bibitem{hamilton_2017_nips}
Hamilton, W., Ying, Z., and Leskovec, J., 2017.
\newblock ``Inductive representation learning on large graphs''.
\newblock In Advances in Neural Information Processing Systems, pp.~1024--1034.

\bibitem{Maas13rectifiernonlinearities}
Maas, A.~L., Hannun, A.~Y., and Ng, A.~Y., 2013.
\newblock ``Rectifier nonlinearities improve neural network acoustic models''.
\newblock In in ICML Workshop on Deep Learning for Audio, Speech and Language
  Processing.

\bibitem{xu2018representation}
Xu, K., Li, C., Tian, Y., Sonobe, T., Kawarabayashi, K.-i., and Jegelka, S.,
  2018.
\newblock ``Representation learning on graphs with jumping knowledge
  networks''.
\newblock In International Conference on Machine Learning, PMLR,
  pp.~5453--5462.

\bibitem{ioffe2015batch}
Ioffe, S., and Szegedy, C., 2015.
\newblock ``Batch normalization: Accelerating deep network training by reducing
  internal covariate shift''.
\newblock In International conference on machine learning, PMLR, pp.~448--456.

\bibitem{he2015delving}
He, K., Zhang, X., Ren, S., and Sun, J., 2015.
\newblock Delving deep into rectifiers: Surpassing human-level performance on
  imagenet classification.

\bibitem{10.5555/1565455.1565462}
Wang, P., 2007.
\newblock ``From nars to a thinking machine''.
\newblock In Proceedings of the 2007 Conference on Advances in Artificial
  General Intelligence: Concepts, Architectures and Algorithms: Proceedings of
  the AGI Workshop 2006, IOS Press, p.~75–93.

\bibitem{ashby_materials_2011}
Ashby, M.~F., 2011.
\newblock {\em Materials selection in mechanical design}, 4th ed~ed.
\newblock Butterworth-Heinemann, Burlington, MA.
\newblock OCLC: ocn639573741.

\bibitem{article}
Cheng, Z., and Ma, Y., 2017.
\newblock ``Explicit function-based design modelling methodology with
  features''.
\newblock {\em Journal of Engineering Design, \textbf{ 28}}, 02, pp.~1--27.

\bibitem{merayo2020prediction}
Merayo, D., Rodr{\'\i}guez-Prieto, A., and Camacho, A.~M., 2020.
\newblock ``Prediction of physical and mechanical properties for metallic
  materials selection using big data and artificial neural networks''.
\newblock {\em IEEE Access, \textbf{ 8}}, pp.~13444--13456.

\bibitem{kingma2017adam}
Kingma, D.~P., and Ba, J., 2017.
\newblock Adam: A method for stochastic optimization.

\bibitem{loshchilov2017sgdr}
Loshchilov, I., and Hutter, F., 2017.
\newblock Sgdr: Stochastic gradient descent with warm restarts.

\bibitem{hochreiter1997long}
Hochreiter, S., and Schmidhuber, J., 1997.
\newblock ``Long short-term memory''.
\newblock {\em Neural computation, \textbf{ 9}}(8), pp.~1735--1780.

\end{thebibliography}

%%%%%%%%%%%%%%%%%%%%%%%%%%%%%%%%%%%%%%%%%%%%%%%%%%%%%%%%%%%%%%%%%%%%%%
\appendix       %%% starting appendix
\begin{appendices}
\onecolumn

\section{Appendix A: Dataset and Feature Distributions}
\hspace{0 pt}
\subsection{List of Features Extracted from the Dataset}
\label{appendix:feature-list}

\begin{table*}[h!]
   \caption{\uppercase{ \footnotesize List of extracted features and encoding methods.}}
    \label{table:assembly-features}
    \centering
    \small
    \begin{tabular}{p{2cm}|p{2.5cm}|p{8cm}|p{1.5cm}|p{1cm}}
    \addlinespace[\aboverulesep] 
    \toprule
        \textbf{Category} & \textbf{Design Feature} & \textbf{Description} & \textbf{Encoding Method} & \textbf{Feature Type}\\
        \midrule
        \textbf{Semantic} & \code{Body name} & Semantic name of the body. & TechNet & Node\\
         & \code{Occurrence name} & Semantic name of the occurrence of the body. & TechNet & Node\\
         \midrule
        \textbf{Geometric} & \code{Body geometry} & Geometry of the body in OBJ format. &  MVCNN & Node\\
         & \code{Body physical \newline properties} & Properties of the body (center of mass, surface area, volume). & Float & Node\\
         & \code{Occurrence  physical properties} & Properties of the occurrence (surface area, volume). & Float  & Node\\
        \midrule
        \textbf{Material} & \code{Material} & The material of the body, treated as ground-truth label. & One-hot  & Node\\
        & \code{Material class} & The hierarchical classification of the material, treated as a user-introduced feature only in the user-guided experiment. From \textit{tier 1} (broad) to \textit{tier 3} (specific). & One-hot  & Node\\
        \midrule
        \textbf{Global}
        & \code{Assembly physical properties} & The physical properties of the overall assembly (center of mass, volume). & Float   & Global\\
        & \code{Assembly geometric properties} & The geometric properties of the overall assembly (total number of edges, faces, loops, shells, vertices). & Integer   & Global\\
        & \code{Design category} & The category specified by the user (automotive, art, electronics, engineering, game, machine design, interior design, product design, robotics, toys, etc.). & One-hot  & Global\\
        & \code{Industry} & The industry specified by the user to describe the assembly (architecture, engineering \& construction; civil infrastructure; media \& entertainment; product design \& manufacturing; other industries). & One-hot  & Global\\
        & \code{Products} & The products used to create the design. & One-hot  & Global\\

        \midrule
        \textbf{Hierarchical} & \code{Hierarchical} & The designer-defined hierarchy, between bodies that share an occurrence. & One-hot  & Edge \\
         & \code{Joints} & Constraints defining the relative pose and degrees of freedom between a pair of occurrences. & One-hot  & Edge \\
         & \code{Contacts} & Faces that are in contact between different bodies. & One-hot  & Edge \\
        \bottomrule
    \end{tabular}

\end{table*}

\newpage
\subsection{Statistics of Material Categories}
\label{appendix:dataset-statistics}

\begin{table}[h!]
\caption{\uppercase{ \footnotesize List of the top-20 material categories, their common names, and counts.}}
\label{table:material-count}
\centering
\begin{tabular}{l|l|l}
\addlinespace[\aboverulesep] 

    \toprule
    \textbf{Material id}       & \textbf{Material name}   & \textbf{Count} \\
    \midrule
    PrismMaterial-018 & Steel                             & 40054 \\
    PrismMaterial-022 & ABS Plastic                       & 2657  \\
    PrismMaterial-002 & Aluminum                          & 2622  \\
    Prism-112         & Plastic - Glossy (Black)          & 2435  \\
    Prism-052         & Gold - Polished                   & 2286  \\
    MaterialInv\_029  & Steel, Mild                       & 1985  \\
    Prism-027         & Aluminum - Polished               & 1954  \\
    Prism-089         & Paint - Enamel Glossy (Black)     & 1593  \\
    Prism-094         & Paint - Enamel Glossy (White)     & 1361  \\
    Prism-047         & Chrome                            & 1262  \\
    Prism-029         & Aluminum - Anodized Glossy (Blue) & 1235  \\
    Prism-113         & Plastic - Matte (Black)           & 1226  \\
    Prism-322         & Bamboo Light - Semigloss          & 1206  \\
    PrismMaterial-017 & Stainless Steel                   & 1110  \\
    Prism-042         & Brass - Polished                  & 1050  \\
    Prism-126         & Plastic - Glossy (White)          & 1042  \\
    Prism-256         & Steel - Satin                     & 1019  \\
    PrismMaterial-003 & Brass                             & 997   \\
    Prism-127         & Plastic - Matte (White)           & 987   \\
    Prism-230         & Iron - Polished                   & 967   \\
    -                 & Other materials                   & 62254 \\
    \midrule
    \multicolumn{2}{l|}{Total} & 131302\\
    \bottomrule
\end{tabular}

\end{table}

\newpage

\subsection{Statistics of Constructed Graphs}
The statistics of the constructed graphs are as follows. Each graph has an average of 24 nodes, a maximum of 821, and a standard deviation of 43.1. Each graph has an average of 40 edges, a maximum of 2336, and a standard deviation of 79.9.

\begin{figure*}[h!]
\centering
\includegraphics[width=18cm]{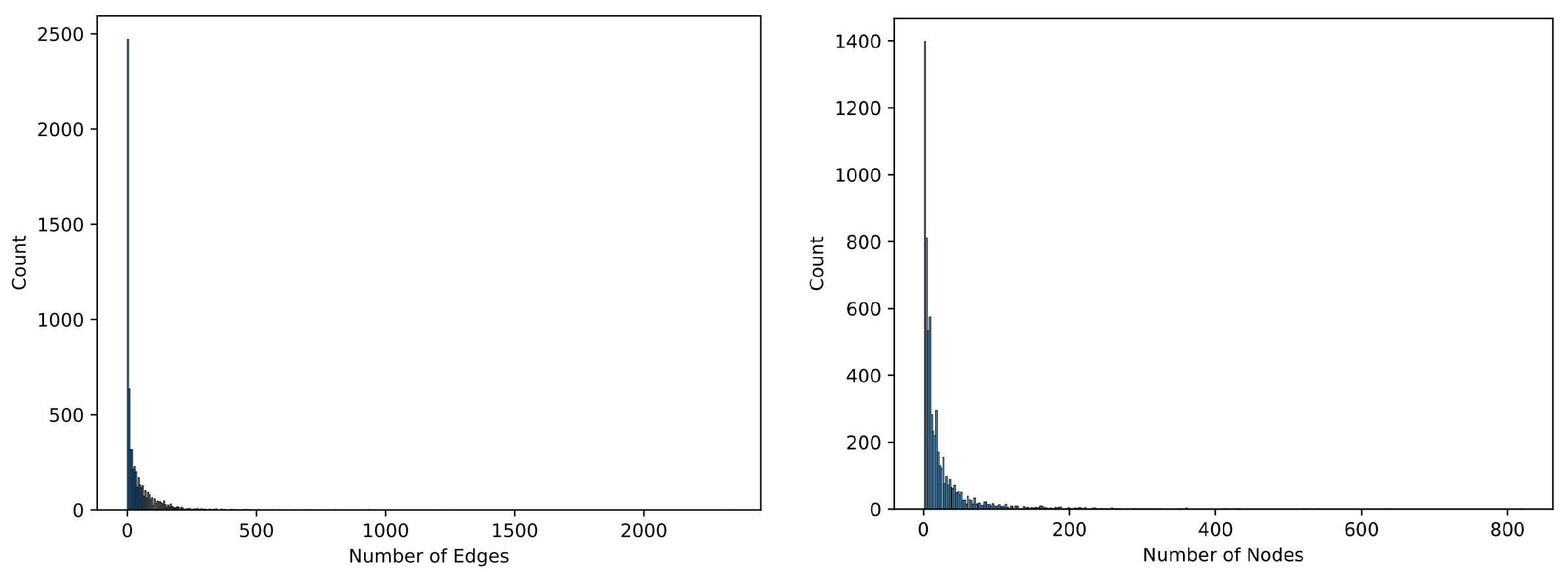}
\caption{ \footnotesize Distribution of Graph Nodes and Edges.}
\label{fig:node_statistics} 
\end{figure*}

\begin{figure*}[h!]
\centering
\includegraphics[width=12cm]{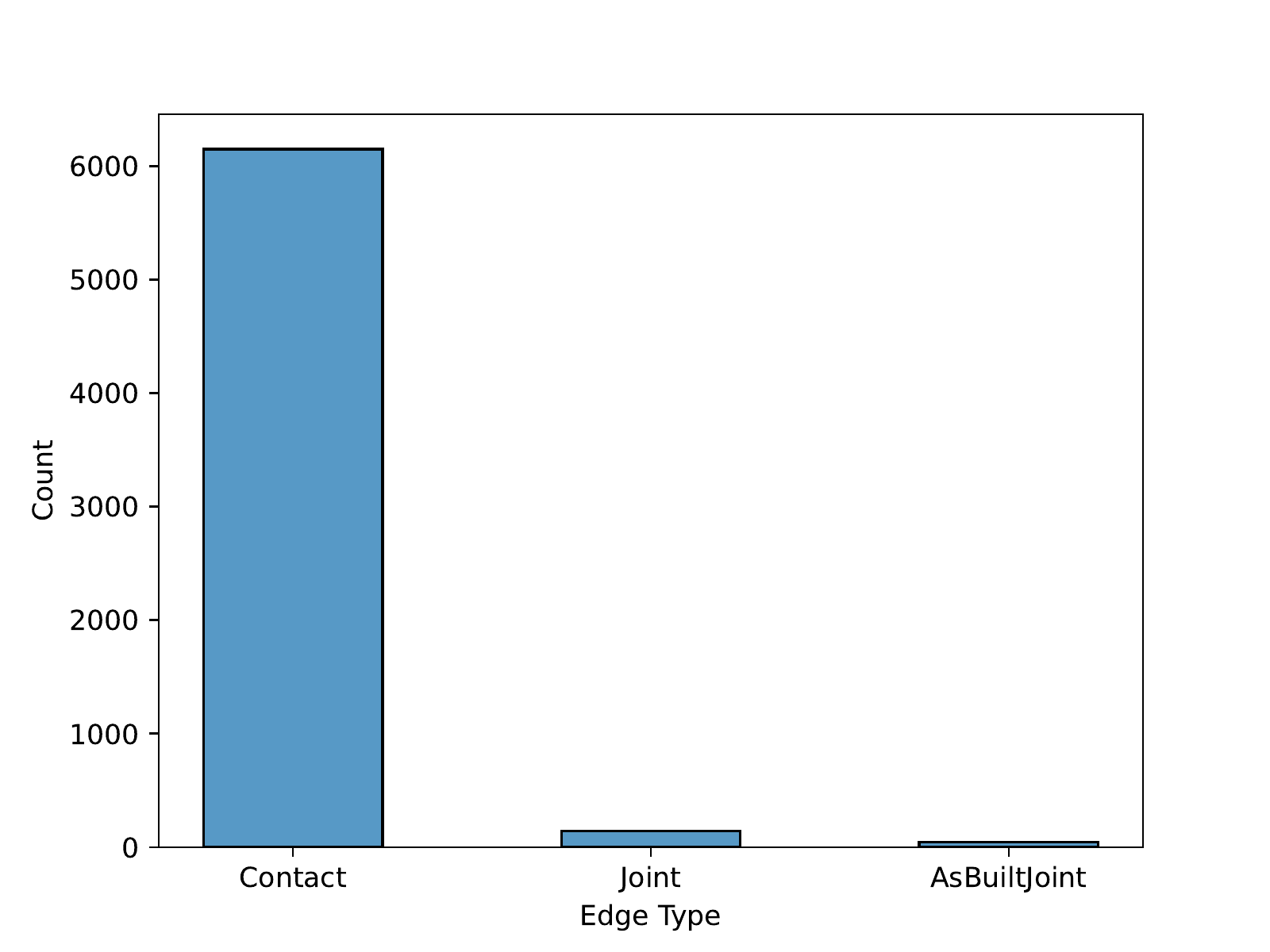}
\caption{ \footnotesize Distribution of Graph Edge Types.}
\label{fig:node_statistics} 
\end{figure*}

\newpage

\section{Appendix B: Additional Explanation of Methods}

\subsection{Feature Extraction} \label{appendix:feature_extraction}
As introduced in Section~\ref{source-of-data}, most features of the assemblies are encapsulated inside their corresponding JSON files and ordered in the format of nested dictionaries. Therefore, we devised the following process to extract the features of each body and their corresponding correlations in the assembly. These extracted features were further encoded following the methodology in Section \ref{sec:feature_engineering} and used for graph construction in Section \ref{sec:graph-construction}.

% \paragraph{Root bodies} A root body is found in the root component of the design as defined by the designer. For each component in the dictionary's \code{root}, we search the bodies within and extract their body features through tracking their UUIDs. These bodies are treated as “root bodies” since they do not have a corresponding occurrence and are given the occurrence name “root” as well as occurrence physical properties of 0. 

\paragraph{Hierarchical bodies} For each occurrence in the dictionary's \code{tree}, we search the bodies within and extract their body features by tracking their UUIDs. After filtering out the bodies that are indicated as invisible, we combine each remaining body feature and their corresponding occurrence features as its final features. Note that the occurrences in “tree” may have a nested structure (i.e., each occurrence may have sub-occurrences) and are therefore hierarchical. To reflect this, we recurrently perform the searching process and assign each body with a hierarchical depth parameter. These features are used later for node creation.

\paragraph{Assembly relationship} For each connection type (\code{joints}, \code{as-built joints}, \code{contacts}) in the JSON file, we search the entities within that contain a body and a corresponding occurrence that are visible. We match these bodies via their UUID to the root and hierarchical bodies we extracted prior to this step and pair them up with their connection type. These features are used later for edge creation.

\subsection{GNN and MVCNN Training Configurations}
\label{appendix:training-parameters}
For the GNN training process, we adopted the Adam optimizer \cite{kingma2017adam} with the learning rate initialized to be 0.001 and a cosine annealing scheduler \cite{loshchilov2017sgdr} with the maximum iteration set to the number of epochs to tune the learning rate for minimizing the loss. After obtaining the embedding from each GNN layer, we apply the Leaky Rectified Linear Unit with a negative slope of 0.2 for non-linear activation. 

\label{appendix:mvcnn-training-parameters}
The MVCNN is trained using a PyTorch implementation and uses the ResNet architecture with a supervised learning regime. A patience factor is used to stop the training process after 20 epochs that see an increase in the validation accuracy, resulting in around 30 training epochs. The models are trained with a batch size of 8, 512 embedding dimensions, 12 views, and $1\mathrm{e}{-4}$ learning rate. 

\subsection{GNN Layer Formulations}\label{sec:gnn_formulation}
As discussed in Section \ref{sec:learning_framework} and Section \ref{sec:general-setup}, we experimented with the GNN layer from the set of $\{SAGE, GAT, GIN, GCN\}$ through a variation of the combination and aggregation functions. After tuning the hyperparameters, we adopted the GraphSAGE neural network, which is particularly suitable for the embedding generation of un-seen nodes on large and diversely structured graphs through sampling and aggregating neighborhood information. The formulation of the parametric combination function $f_\theta(.)$ and the aggregation function $g_\phi(.)$ for the GraphSAGE network at layer $k$ are as follows. 

\paragraph{GraphSAGE \cite{hamilton_2017_nips}} The aggregation and combination processes for GraphSAGE are formulated in Equation \ref{equation:sage_formulae}. For the first equation that formulates the aggregation process, $a_{v}^{(k)}$ indicates the aggregated feature for node $v$ at layer $k$, $LSTM$ indicates a Long Short-Term Memory aggregator \cite{hochreiter1997long}, $h_{u}^{k-1}$ indicates the node feature for node $u$ at layer $(k-1)$, and $N(v)$ indicates the set of nodes neighboring to node $v$. For the second equation that formulates the combination process, $ReLu$ indicates a Rectified Linear Unit, $||$ indicates matrix concatenation operation, and $W^k$ represents a weight matrix for layer $k$ that is learnable. GraphSAGE allows for the efficient generation of embeddings for un-seen nodes through sampling and aggregating neighborhood information and is suitable for representation learning on large graphs.

\begin{equation}\label{equation:sage_formulae}
    a_{v}^{(k)} = LSTM(\{h_{u}^{(k-1)}, u \in N(v)\}), \quad h_{v}^{(k)} = ReLU(W^{(k)}[h_v^{(k-1)} || a_v^{(k)}])
\end{equation}

% % GAT: https://arxiv.org/pdf/1710.10903.pdf
% \paragraph{Graph Attention Network (GAT) \cite{velickovic_2018_iclr}} The aggregation and combination processes for GAT are formulated in Equation \ref{equation:gat_formulation}. For the first equation, $\alpha_{ij}^{(k)}$ indicates the \textit{attention coefficients} that represents the \textit{importance} of node $j$'s feature to node $i$, $\sigma$ indicates the Leaky Rectified Linear Unit (LeakyReLU), $W$ and $a$ are the weight matrix and vector that are learnable. Note that $j \in \{i\} \cup N(i)$ are the first-order neighbors of node $i$. 

% \begin{equation}\label{equation:gat_formulation}
%     \alpha_{ij}^{(k)} = \frac{exp(\sigma(a^{T}[Wh_i^{(k-1)}||Wh_{j}^{(k-1)}]))}{\sum_{l\in N(i)}exp(\sigma(a^{T}[Wh_i^{(k-1)}||Wh_{l}^{(k-1)}]))}, \quad h_i^{(k)} = ReLU( \sum_{j \in N(i)} \alpha_{ij}^{(k)} W h_j^{(k-1)}) 
% \end{equation}
% % j\in\{i\}\cup N(i)}^{(k-1)

\end{appendices}
\end{document}